\documentclass[hf
]{ceurart}

\begin{document}

\copyrightyear{2022}
\copyrightclause{Copyright for this paper by its authors.
  Use permitted under Creative Commons License Attribution 4.0
  International (CC BY 4.0).}

\conference{PMAI@IJCAI22: International IJCAI Workshop on
Process Management in the AI era, July 23, 2022, Vienna, Austria}

\title{Prescriptive Process Monitoring in Intelligent Process Automation with Chatbot Orchestration}

\author[1]{Sergey Zeltyn}[%
email=sergeyz@il.ibm.com]

\author[1]{Segev Shlomov}[%
email=segev.shlomov1@ibm.com]

\author[1]{Avi Yaeli}[%
email=aviy@il.ibm.com]

\author[1]{Alon Oved}[%
email=alon.oved@il.ibm.com]

\address[1]{IBM Research - Israel,  Haifa University Campus, Mount Carmel Haifa, 3498825, Israel}


\begin{abstract}
Business processes that involve AI-powered automation have been gaining importance and market share in recent years. These business processes combine the characteristics of classical business process management, goal-driven chatbots, conversational recommendation systems, and robotic process automation. In the new context, prescriptive process monitoring demands innovative approaches. Unfortunately, data logs from these new processes are still not available in the public domain. We describe the  main challenges in this new domain and introduce a synthesized dataset that is based on an actual use case of intelligent process automation with chatbot orchestration. Using this dataset, we demonstrate crowd-wisdom and goal-driven approaches to prescriptive process monitoring.  
\end{abstract}


\maketitle
\thispagestyle{empty}

\section{Introduction}
\label{sec:introduction}
In recent years, the term intelligent process automation (IPA) has been used to describe a new category of digital workers. IPA combines a mix of technologies from robotic process automation (RPA) \cite{aalst2018robotic}, natural language processing (NLP), machine learning (ML), artificial intelligence (AI), and traditional digital business process automation and integration platforms.  By combining these technologies, IPA promises to support a wider and more complex range of process automation needs, and enable non-professional developers to build, deploy, and engage more naturally with digital workers to perform their tasks.  Overall, IPA is expected to become a key component for orchestrating human-to-bot work and is already being adopted across a variety of industries and business areas such as finance, human resources, operations, and sales.


An initial survey of IPA vision and characteristics was done  by Chakraborti et al. \cite{chakraborti2020robotic}. One of the key characteristics of IPA is a conversational interface, since this represents a very natural way for humans to interact and collaborate with bots as part of the automation process.  Another key component is the ability to dynamically orchestrate RPA and other types of automation based on dynamic and contextual user needs.  For example, Rizk et al.  \cite{rizk2020conversational} describe a conversational IPA platform that uses AI planning to orchestrate conversational dialogues and automation sequences based on user utterances and a pre-populated catalog of automation skills.


As IPA platforms become more powerful, we can expect a new generation of conversation-oriented digital employees that can handle complex and dynamic processes. The duration of these processes can span from seconds and minutes to weeks and months.


\vspace{0.2cm}
\textbf{Prescriptive process monitoring (PPM) }techniques are used to improve business processes by triggering interventions at runtime to optimize the process towards a goal such as a key performance indicator (KPI). Traditionally, PPM looks at case ID, activity timing and sequences, resources, and business attributes to predict the progress towards the goal and introduce interventions.   Because IPA has many unique elements when compared to traditional BPM, PPM requires adaptations to support prescriptive tasks in IPA. 


One scenario of these adaptations occurs when there is a need to prescribe a recommendation in the form of a human-to-bot interaction, such as a button or textual utterance. There are four possible situations to consider:  First, the discovery of skills and utterances for new users who are not yet aware of what the bot can do and which utterances will trigger those skills. Second, a recovery from interaction failure.  Occasionally the bot will not understand the user or an error may occur that will cause the interaction to derail from its original context.  Third, goal-oriented actions that help or remind the user to perform some activity in order to achieve the process goals.  Fourth, the wisdom of the crowd and personalization that can recommend possible actions based on the behavior of other users, or on personal preferences.

Aside from the type of prescription, there are additional characteristics that are different from classical PPM, such as how process and session identifiers in human-to-bot interaction map to the concept of case ID, how disambiguation and errors are modeled as activities, how to treat or leverage human feedback, and the types of intervention that can achieve user engagement with the bot.


IPA is a relatively new domain and real-world deployments are not available for researchers due to confidentiality. Therefore, we believe that a synthetic dataset, inspired by actual IPA deployment, will be of value in terms of understanding the data model, testing existing algorithms, and developing new ones that can later be validated on real-world datasets.


{\bf The contribution} of this paper is twofold. First, we introduce and share a first-of-a-kind synthetic dataset based on a real-world use case for an HR Management Incentive Program (MIP). This dataset presents a new type of data from IPA-based processes. Second, we demonstrate an implementation of crowd-wisdom and goal-oriented prescription tasks for this new dataset and explain the IPA-related adaptations to traditional algorithms.

The remaining part of the paper is organized as follows. Section \ref{sec:related_art} discusses the related art. Section \ref{sec:mip_dataset} introduces the MIP dataset. Specifically, Section \ref{subsec:dataset_story} explains its business use case, Section \ref{subsec:dataset_design} presents parameters of the dataset simulation and Section \ref{subsec:dataset_description} provides the dataset schema.  The following sections apply two approaches for  prescriptive process monitoring to the MIP dataset. Section \ref{sec:crowd_wisdom} describes the crowd-wisdom approach based on the prediction of the next activity and Section \ref{sec:goal_driven} explains the goal-driven approach based on lateness prediction. Finally, Section \ref{sec:summary} summarizes the paper and future research challenges.

\section{Related Art}
\label{sec:related_art}
The basic goal of prescriptive process monitoring (PPM) is to optimize a process at run-time \cite{kubrak2021prescriptive}. The three main PPM methods are predictions (e.g., next activity \cite{de2020design,weinzierl2020recommender}), corrective actions, and resource optimization (e.g., which resource should perform the next task \cite{wibisono2015fly}). Over the years, many techniques have been developed using classical machine learning techniques \cite{francescomarino2018predictive} and deep learning algorithms \cite{neu2021prediction}.

The recent emergence of trustworthiness in AI systems, has brought explainability and causal papers to the PM domain. Galanti et al.  \cite{galanti2020explainable} showed how explanations can be given in the field of predictive business process monitoring by using Shapley values to obtain explanations of KPI predictions, such as remaining time and activity execution. Bozorgi et al. \cite{bozorgi2020causal} applied causal machine learning techniques to the BPM domain and explored which action should be applied to yield the highest causal effect on the business process outcome. Metzger et al. \cite{metzger2020reinforcement} used a reinforcement learning technique in prescription. They tackled the trade-off of earlier predictions, which leave more time for adaptations but exhibit lower accuracy. The authors described when to trigger proactive process adaptations with online reinforcement learning.

Our paper is also related to recommendations for goal-driven chatbots \cite{lipton2018bbq} and conversational recommendation systems, where many papers suggest to use reinforcement learning techniques \cite{afsar2021recommender,lipton2018bbq, metzger2020reinforcement}. The chatbot research studies systems that are designed by tools such as IBM Watson Assistant,  Google Dialog, and Microsoft’s Cortana. In the conversational recommendation systems, users might ask questions about the recommendations and provide feedback.  In a recent paper, Weinzierl et al. \cite{weinzierl2020recommender} used BPM prediction techniques in recommender systems. They modeled a sequence of user clicks as a process and used an NLP-based process embedding to recommend the next best click. The paper also argues the importance of ``crowd knowledge" for providing good recommendations. 

\vspace{0.2cm}
\textbf{Datasets:} There are many open datasets that are process-based. Examples include the ones created for the BPI 2011-2020 challenges \cite{dongen2020datasets}. They vary in size, domain (e.g., incidents, loans and complaints), and complexity. Other related datasets come from the dialogue domain. Some of them describe human-to-human interaction \cite{lowe2015ubuntu, AirDialogue} while the others describe human-to-bot \cite{logacheva2018convai} interaction. These datasets contain many different tasks, such as intent prediction \cite{mehri2020dialoglue}, slot filling \cite{miller2017parlai}, dialogue state tracking \cite{rastogi2020schema}, and dialog act \cite{budzianowski2018multiwoz,eric2019multiwoz}. While some of these datasets satisfy part of the new domain properties (see also \cite{dumas2022augmented}), none of them satisfy all of them; that is, a dataset that is process-based, contains both session ID and case ID, and consists of multi-person interactions.

\section{MIP Dataset: Use Case and Description}
\label{sec:mip_dataset}

Datasets from IPA systems with chatbot orchestration are still unavailable in the public domain. Thus, we strive to partially close this gap by providing a synthesized dataset based on a real-life use case. Specifically, the use case focuses on human resource (HR) automation systems, which constitute an important application domain for the new technology. Guenole and Feinzig  \cite{IBM_HR_AI_Transformation} summarize the IBM HR business case with automation IPA systems being used for recruiting, onboarding new employees, career coaching, personalized learning, and other important tasks.

As a source of inspiration for our dataset, we used an internal IBM application in the domain of compensation and promotion. The implementation is based on Watson Orchestrator\footnote{https://www.ibm.com/cloud/automation/watson-orchestrate}. This use case satisfies several basic prerequisites for an IPA dataset. It is based on a multi-person business process, contains chatbot conversational interactions, and gives rise to two embedded process identifiers: case ID defines an instance of a business process and session ID defines a chatbot conversation instance.

\subsection{The MIP Dataset Use Case}

\label{subsec:dataset_story}

%
We consider a large software engineering organization. The Management Incentive Program (MIP) process is run in the organization twice a year. The goal of the process is to determine which employees will get a salary increase. 

%
There exist two roles in the MIP process: first-line team leaders and second-line department managers. The team leaders initiate a case instant to select team employees who are eligible to participate in the program. Then they provide the names of nominated candidates to the department managers. For each candidate, a department manager decides whether the candidate nomination should be approved, approved with correction (e.g., the amount of salary increase is changed), or rejected. Then, the department manager submits the final decision to the HR system, completing the case instance for a specific team. In summary, each case instance consists of two sequential tasks, nomination and approval, which are performed by two different users.

%
In order to make informed decisions, the team leaders and department managers go over a number of reports that contain different employee performance metrics and summarize employee activities and feedback. The list of 20 available reports is provided in Table \ref{table:reports}. The table also contains probabilities that a user will look at the reports at least once during the process. These reports differ in their importance and the users view them in a random order with different frequencies. MIP criteria and yearly assessment reports are always viewed by the team leaders before initiating nomination actions. We also observe from Table \ref{table:reports} that the department managers view reports with a lower frequency than the team leaders. 
\begin{table}
\caption{List of reports}
\label{table:reports}
\begin{center}
\begin{tabular}{lcccc}
\toprule
Report &  Viewing by team leader, \% &  Viewing by department manager, \%  \\
\midrule 
MIP criteria  &  100  & 62.6   \\
Yearly assessments  &  100  & 50.8  \\
Project assessments  & 91.1 & 56.7  \\
Learning activities  & 87.2 & 47.0 \\
Client feedback      & 86.5 & 50.7    \\ 
Internal feedback    & 89.4 & 51.8  \\
Compensation report  & 91.6 & 52.7    \\
MIP history          & 91.9 & 56.7     \\ 
Overtime             & 55.9 & 30.1    \\
Innovation and patents   &  70.9  &  35.5    \\
Product defects   & 36.3  & 15.1    \\ 
Sprints velocity   & 32.9    & 17.0    \\
Bugs fixed    &  38.6  & 15.2     \\
Pull requests    & 15.2  & 12.9    \\ 
Features shipped   & 30.3   & 26.1    \\
Defects repair time    & 17.1    & 10.8     \\
Lead time  & 27.3  & 17.1     \\ 
Project costs   &  51.1  &  27.2  \\
Code churn    & 31.0   & 27.1     \\
Absence    & 34.1  & 18.2     \\ 
\bottomrule
\end{tabular} 
\end{center}
\end{table}
The activities that the team leaders and the department managers can perform are as follows. The team leader can: view report, add nomination, view nomination, submit nomination, and provide candidate name. The department manager can: view report, select candidate name, review nominated candidate, approve nomination, approve nomination with correction, reject nomination, and submit final nominations. We assume that most of these activities are initiated via free text chat. The system recognizes a specific intent of a user and performs the activity that corresponds to this intent. Two name selection activities are performed via a slot-filling mechanism that is frequently used by the goal-driven chatbots.
 
The use of free text to trigger activities gives rise to the two additional scenarios. First, sometimes the IPA system cannot detect a user intent with sufficient confidence. In this case, the user utterance is followed by a {\em fallback}: a user is asked to provide additional input. Second, a user utterance can potentially correspond to more than one intent. A standard solution for this problem is {\em disambiguation} - the  widely used prescriptive technique in the chatbot domain. The user is asked to select between several competing intents via the corresponding buttons. In the MIP dataset, we consider four disambiguation scenarios that are shown in Table \ref{table:disambiguation}.
\begin{table}
\caption{List of disambiguation scenarios with examples}
\label{table:disambiguation}
\begin{center}
\begin{tabular}{lcccc}
\toprule
Activities &  Example  \\
\midrule 
view client feedback report / internal feedback report  &  show feedback report   \\
view project assessment / project cost report &  view project data   \\
view MIP criteria report / MIP history report &  MIP data   \\
view product defects report / defects repair time report & I need defects report \\
\bottomrule
\end{tabular} 
\end{center}
\end{table}
Like many business processes, the MIP process has time constraints that include a regular and a ``hard stop'' deadline.  It is technically possible but undesirable to violate a regular deadline. On the date of a ``hard stop'' deadline, process participants are forced to complete the MIP process within several hours.  

Finally, we assume that both team leaders and department managers are divided into two groups with different statistical properties: struggling users and successful users. Struggling users have, on average, larger intervals between conversation sessions, a higher number of fallbacks, and a higher probability of abandoning a session without successful completion of the task. Naturally, struggling users have a tendency to push the business process over the deadline.

\subsection{Parameters and Statistical Properties of the MIP Dataset}  

\label{subsec:dataset_design}

In this section, we describe some deterministic and statistical properties of the MIP dataset simulation. We assume that the process starts on Monday, Mar 7, 2022 and it is desirable to complete it until the regular deadline: end of Monday, Mar 28, 2022. A second ``hard stop'' deadline takes place on Monday, Apr 11, 2022. Users can interact with the IPA system during a Monday to Friday working week, with working hours between 8am and 5pm, although sometimes conversation sessions take place later in the evening. 
   
There are 250 department managers, and 4 first-line team leaders under each department manager. Overall, there are $250 \cdot 4 = 1,000$ cases of the process, which should provide a sufficiently large data sample for the analysis in Sections \ref{sec:crowd_wisdom} and \ref{sec:goal_driven}. On average, there are 10 employees in each team. The MIP nomination rate is 20\%. 

We assume that the struggling users constitute 1/3 of team leaders and department managers. The main statistical characteristics of successful and struggling users are presented in Table \ref{table:users}.

\begin{table}
\caption{Performance characteristics per role}
\label{table:users}
\begin{center}
\begin{tabular}{lcccc}
\toprule
User type &  Average interval between & Average number  & Probability of fallback  \\
       &  sessions (working days) & of conversations  &  per utterance  \\
\midrule 
team leader, successful  &  1.5   & 2 & 0.05     \\
team leader, struggling  &  4   & 5.5  &  0.3 \\
dep.\ manager, successful  &  1  &  1.5 & 0.05 \\
dep.\ manager, struggling  &  2  & 3.5  & 0.3   \\
\bottomrule
\end{tabular} 
\end{center}
\end{table}
To generate user free-text we applied  Lambada \cite{ateret2020lambada} methodology. For each intent, we used a small manually prepared seed of utterances that was enriched via the Lambada algorithm. 

\subsection{Dataset Schema}  

\label{subsec:dataset_description}

\begin{table}\label{tbl:dataset_description}
\caption{MIP column descriptions}
\label{table:columns}
\begin{center}
\begin{tabular}{l l l l }
\toprule
Feature name & Description  & Example  \\
%
\midrule
case\_id            & corresponds to the MIP process per team               & 1         \\
session\_id         & conversation session id                               & M7vkTk2f537I      \\
role                & role of the user                                  & team leader       \\
user\_id            & id of the user                                    & Robert North       \\
timestamp           & timestamp of a user utterance                                & 2022-03-17T11:20:21      \\
turn                & current turn in a conversation                                     & 2      \\
activity           & IPA activity  triggered by the user utterance   & report\_lead\_time        \\
user utterance      & user utterance that triggers chatbot and system activities                   & view lead time table      \\
chatbot response    &  chatbot response to the user utterance                       & Lead Time Report      \\
intent              &   intent triggered by the user utterance           & report\_lead\_time      \\
intent\_confidence  & conversation engine confidence in the user intent        & 0.898      \\
entity              & appears when a user is engaged in slot-filling         &   Troy Donovan    \\
entity\_confidence  &   conversation engine confidence in user entity & 1.0       \\
score               & IPA system orchestrator score based on the intent     &  0.862 \\
                    & confidence and other system characteristics           & \\
expecting\_response & indicates if a user is expected to answer the chatbot question & False &\\

\bottomrule
\end{tabular} 
\end{center}
\end{table}

The MIP dataset is provided in {\em csv} format\footnote{https://github.com/Sergey-Zeltyn/MIP-dataset}.
Table \ref{table:columns} presents the names of the dataset columns with brief descriptions and examples. 

Each row of the file corresponds to the turn of a conversation with a chatbot. During all turns, except those corresponding to the chatbot welcome message, a user utterance triggers an activity in the IPA system. For some turns, the system recognizes a user intent, in which case the name of the activity coincides with the intent name. Additional activities are responsible for slot filling and user utterances with unrecognized intent (fallbacks and disambiguations).

We assumed that the activities are orchestrated using an agent orchestration concept similar to the one used by Rizk et al.\ \cite{rizk2020conversational}. The score concept we used for activity selection by an IPA system was also introduced in \cite{rizk2020conversational}.  Since chatbot responses are typically not used in the prescription methods, brief stub utterances (for example, ``Welcome Message'') replace them in the dataset.  

\section{Crowd-Wisdom Prescriptions}
\label{sec:crowd_wisdom}
%
New users of IPA systems are frequently not fully aware of the actions that they can perform. They need a straightforward and intuitive way to flatten the learning curve. Crowd-wisdom methods help achieve this goal. Experienced users can implicitly guide inexperienced users to ``happy paths''. For example, in the MIP use case, the path statistics for advanced users can be used to recommend the most important performance reports. 

%
Prediction of the next activity or the next several activities is the key stage of the crowd-wisdom approach. Such prescription systems typically present several options for the user's next actions. As a result, the underlying predictions should be probabilistic and not single-activity ones. At the same time, undesirable activities, such as fallback or disambiguation, should not be mentioned in recommendations, even if they are performed frequently by other users. 

In this section, we focus on predicting the next activity for the MIP dataset, while emphasizing several different feature generation approaches. There are 36 activities overall: 
\begin{itemize}
\item viewing of 20 reports, 4 disambiguation activities, and fallback for both roles;
\item candidate nomination, candidate name selection, viewing  nomination, and submitting nomination for the team leaders; 
\item reviewing nomination, nominee name selection, selecting one of three possible decisions, and final submission for the department managers.
\item session end should be considered an additional activity in the prediction problem. 
\end{itemize}
For the prediction task, we do not filter out undesirable activities and compare different prediction methods based on the overall goodness-of-fit.

%
Feature generation is especially important in IPA processes that involve chatbots, where conversation sessions constitute sub-processes with special characteristics. For each prediction technique, we consider two feature generation dilemmas. The first dilemma is how to extract the process features: is information on the previous turn sufficient for prediction or should it be complemented by process-aware features? In the non-process-aware (npa) approach, features include the previous activity and several attributes of the previous conversation turn. The list of the attributes includes role, intent confidence, score, expecting\_response, and the number of turns in a session. The process-aware (pa) approach adds process path statistics to the feature vector. In our case, this path statistics includes the number of occurrences of each activity during the current session until the current conversation turn.

The second, more subtle dilemma, is related to the session definition. Should it be based on conversation (option \textit{conv} in Table \ref{table:crowd_comparison})? Or should we base it on the overall path of a user, even if a user was engaged in several conversations on different days (\textit{user} option in Table \ref{table:crowd_comparison})?  This is an important question since a user can behave differently over different sessions or forget details of the previous ones. In the conv process-aware setting, we count activities from the start of conversations and add a sequential conversation number of a specific user to the feature space. In the user process-aware setting,  we count activities from the first user login into the system. Finally, we combine the two approaches and use the union of features from the two session definitions  (\textit{conv+user} option in Table \ref{table:crowd_comparison}). In Table \ref{table:crowd_comparison}, we compare the three process-aware approaches described above and an implementation of a non-process-aware approach.

%
We implemented three prediction techniques: logistic regression, CatBoost, and XGBoost. Our goodness-of-fit metrics included accuracy, weighted Top-3 recall, average Top-3 recall, weighted $F_1$ score, and average $F_1$ score. We computed the averages between 35 prediction classes, while leaving out the ``end'' class since it is defined differently for different session definitions. Weighted averages were weighted by volume in the testing sets. The Top-3 recall can be interpreted as a fraction of samples where the actual activity belongs to the Top-3 predicted activities, ordered by predicted probabilities. This metric is important because crowd-wisdom recommendations typically provide several alternatives. We performed 5-fold cross-validation and averaged metrics over 5 testing datasets.  

\begin{table}
\caption{Goodness of-fit for different methods of next activity prediction}
\label{table:crowd_comparison}
\begin{center}
\begin{tabular}{lccccc}
\toprule
Prediction method & Weighted     & Average        & Weighted     & Average      & Accuracy    \\
                  & Top-3 Recall & Top-3 Recall   & $F_1$-score & $F_1$-score &  \\
%
\midrule
Logistic Regression, npa, conv     & 0.519 & 0.412 & 0.230 & 0.176 & 0.302 \\
Logistic Regression, pa, conv      & 0.588 & 0.490 & 0.293 & 0.241 & 0.351 \\
Logistic Regression, pa, user      & 0.565 & 0.457 & 0.287 & 0.233 & 0.348 \\
Logistic Regression, pa, conv+user & 0.596 & 0.497 & 0.306 & 0.251 & 0.361 \\
CatBoost, npa, conv     & 0.518 & 0.414 & 0.243  & 0.196  & 0.313 \\
CatBoost, pa, conv      & 0.586 & 0.470 & 0.298  & 0.238  & 0.367 \\
CatBoost, pa, user      & 0.566 & 0.445 & 0.283  & 0.214  & 0.355 \\
CatBoost, pa, conv+user & 0.596 & 0.477 & 0.311  & 0.244  & 0.380 \\
XGBoost, npa, conv      & 0.518 & 0.415 & 0.245  & 0.200  & 0.311 \\
XGBoost, pa, conv       & 0.605 & 0.500 & 0.317  & 0.265  & 0.375 \\
XGBoost, pa, user       & 0.580 & 0.469 & 0.300  & 0.241  & 0.362 \\
XGBoost, pa, conv+user  & {\bf 0.616} & {\bf 0.511} & {\bf 0.330}  & {\bf 0.270}  & {\bf 0.390} \\ 
\bottomrule
\end{tabular} 
\end{center}
\end{table}
%
%
Table \ref{table:crowd_comparison} indicates that process-aware features significantly improve the goodness-of-fit for all prediction settings under  consideration.  XGBoost, which uses the union of features from the two session definitions, implies the best goodness-of-fit. The conversation-based session definition shows better results than the user-based definition for all prediction techniques. Given the large number of classes overall and the fact that 20 performance reports had very significant variance in their sequences, the accuracy and Top-3 recall numbers are satisfactory. For example, a random class selection would imply approximately 0.03 accuracy.

\section{Goal-Driven Prescriptive Process Monitoring}
\label{sec:goal_driven}

The crowd-wisdom approach, presented in Section \ref{sec:crowd_wisdom}, is a useful one but, in many circumstances, should be complemented by a goal-driven prescription setting. Processes in the IPA domain can have goals related to process time, cost, quality, and outcomes. One of mainstream approaches in business process management uses a two-step method \cite{teinemaa2018alarm}. First, prediction is performed for the current process instance with respect to process goals. Second, a corrective action is implemented for the instances with unsatisfactory predictions.     

For the MIP dataset, we address the binary prediction problem of the Mar 28, 2022 deadline violation. In the MIP dataset, 32\% of the process instances violated the deadline. We do not explicitly simulate a corrective action assuming that it could be a reminder mail to a user.

\begin{table}
\caption{Goodness of-fit for different methods of lateness prediction}
\label{table:lateness_comparison}
\begin{center}
\begin{tabular}{lcccc}
\toprule
Prediction method  & Precision & Recall  & $F_1$-score & Accuracy \\
\midrule
Logistic Regression, npa, conv     &  0.934  & 0.879  & 0.906  & 0.926 \\
Logistic Regression, pa, conv      &  0.933  & 0.885  & 0.908  & 0.927 \\
Logistic Regression, pa, user      &  0.938  & 0.897  & 0.917  & 0.934 \\
Logistic Regression, pa, conv+user &  0.934  & 0.901  & 0.917  & 0.934 \\
CatBoost, npa, conv                &  0.941  & 0.892  & 0.916  & 0.933 \\
CatBoost, pa, conv                 &  0.939  & 0.914  & 0.926  & 0.941 \\
CatBoost, pa, user                 &  {\bf 0.940}  & {\bf 0.930}  & {\bf 0.936}  & {\bf 0.947} \\
CatBoost, pa, conv+user            &  0.939  & 0.927  & 0.933  & 0.946 \\
XGBoost, npa, conv                 &  0.936  & 0.889  & 0.911 & 0.929 \\
XGBoost, pa, conv                  &  0.936  & 0.909  & 0.922 & 0.938 \\
XGBoost, pa, user                  &  0.938  & 0.928  & 0.933 & 0.945 \\
XGBoost, pa, conv+user             &  0.938  & 0.925  & 0.931 & 0.944 \\
\bottomrule
\end{tabular} 
\end{center}
\end{table}
We apply the same feature generation methods and prediction techniques as in Section \ref{sec:crowd_wisdom} with a single change:  the timestamp of the current turn is added to the features in all settings. The timestamp is transformed into the number of working days since the start of the process.  We use the standard metrics for binary classification problems to compare prediction methods.  

%
Table \ref{table:lateness_comparison} summarizes the prediction results. The process-aware approach performs better than the non-process-aware one. In contrast to Section \ref{sec:crowd_wisdom}, the definition of our user-based session is preferable to the conversation-based method and implies a better balance between precision and recall. Applying a combination of the two approaches does not improve the goodness-of-fit. Although the results for the three prediction techniques are relatively close, CatBoost produces the best predictions.  

An analysis of the feature importance values for XGBoost provides insights into the results above. In addition to the user role and timestamp, which clearly affect lateness predictions, the number of fallbacks during the session is also in the Top-3 features based on importance. The possible reason is that struggling users have both a significant number of fallbacks and a high probability of being late for the deadline. It is reasonable to assume that the user-based session definition provides more reliable fallback statistics than a conversation-based one. In addition, observation on the number of fallbacks demonstrates that a language-based conversation feature can be important for long-term process prediction: users with many fallbacks can be identified as risk-prone ones at an early stage of the process and provided with assistance.

\section{Summary}
\label{sec:summary}

We presented the emerging domain of intelligent process automation bots with a chat interface.  We highlighted unique aspects of PPM for this new domain, such as the type of prescriptions and data model mapping, which require adaptations of traditional prescriptive approaches to BPM.  We further introduced HR MIP - a synthetic dataset that is inspired by real-world IPA deployment.  This first-of-a-kind dataset can be used by researchers to develop and test new algorithms in this domain.  In addition, we presented an implementation of crowd-wisdom and goal-oriented prescriptive tasks and used it to illustrate the necessary adaptations to data mapping and feature generation steps. We hope that these contributions and the availability of an open dataset will be of value to other researchers in the community.

Prescriptions in intelligent process automation with a chat interface include many additional challenges for future research.  The dynamic nature of AI-based digital employees and human-to-bot interaction may entail continuous concept drift.  In this setting, explore-exploit techniques such as reinforcement learning could be applied  to leverage  implicit and explicit user feedback.  Another challenge is how to deal with more complex utterances, e.g., that are handled by an AI planner to dynamically orchestrate robotic process management tasks.  Such use cases may require deep learning, NLP, and program synthesis approaches to map from user utterances to activities and then back to textual recommendations. We plan to address some of these challenges in the next version of the HR IPA system that inspired the MIP use case. 
 
 \acknowledgments{The authors thank Slobodan Radenković, Tomáš Bene, Yara Rizk, Vatche Isahagian, and Vinod Muthusamy for fruitful discussions on methodology and use cases in intelligent process automation.}

%
%
%


\begingroup
\raggedright
\bibliography{references}
\endgroup

\end{document}